\documentclass{article}

\usepackage[final]{custom_2021}

\usepackage{natbib}
\usepackage{graphicx} 
\usepackage{subcaption} 

\usepackage[utf8]{inputenc} 
\usepackage[T1]{fontenc}    
\usepackage{hyperref}       
\usepackage{url}            
\usepackage{booktabs}       
\usepackage{amsfonts}       
\usepackage{nicefrac}       
\usepackage{microtype}      
\usepackage{xcolor}         
\usepackage{amsmath}
\usepackage{amssymb}
\usepackage{amsfonts}

\title{Sampling To Improve Predictions For Underrepresented Observations In Imbalanced Data
}

\author{%
  Rune D. Kjærsgaard \\
  DTU Compute\\
  \texttt{rdokj@dtu.dk} \\
   \And
   Manja G. Grønberg \\
   DTU Compute\\
   \texttt{mgegr@dtu.dk} \\
   \And
   Line K. H. Clemmensen \\
   DTU Compute\\
   \texttt{lkhc@dtu.dk} \\
}

\begin{document}

\maketitle

\begin{abstract}
Data imbalance is common in production data, where controlled production settings require data to fall within a narrow range of variation and data are collected with quality assessment in mind, rather than data analytic insights. This imbalance negatively impacts the predictive performance of models on underrepresented observations. We propose sampling to adjust for this imbalance with the goal of improving the performance of models trained on historical production data. We investigate the use of three sampling approaches to adjust for imbalance. The goal is to downsample the covariates in the training data and subsequently fit a regression model. We investigate how the predictive power of the model changes when using either the sampled or the original data for training. We apply our methods on a large biopharmaceutical manufacturing data set from an advanced simulation of penicillin production and find that fitting a model using the sampled data gives a small reduction in the overall predictive performance, but yields a systematically better performance on underrepresented observations. In addition, the results emphasize the need for alternative, fair, and balanced model evaluations.

\end{abstract}

\section{Introduction}
\label{sec:intro}
Production data is often gathered under very controlled settings, driven by a requirement of the data to fall within a specified range of variation, and experiments are often expensive leaving data insights to be derived from the available historical data. For this reason, production data commonly exhibit low variation expressed by most of the data lying in high-density areas with only few data points falling outside these areas. This is called imbalanced data and has been studied extensively for categorical targets (\cite{haixiang2017learning}, \cite{krawczyk2016learning}), but only sparsely for continuous targets (\cite{branco2017smogn}, \cite{branco2019pre}). Previous works consider the imbalance to be caused by the target, where we on the other hand consider the imbalance mainly driven by the input variables. The premature ideas for this research were developed in \cite{posterENBIS}.

Imbalance in the response variables is often handled through data-level approaches like over- or undersampling the classes, or through algorithm-level approaches like e.g. class priors, or by use of a hybrid of these (\cite{krawczyk2016learning}, \cite{johnson2019}). Here, we extend the data-level line of thought to consider sampling with respect to the input space. The assumption is that a balanced representation of the input space gives better inference for underrepresented parts of the input space. Thus, we propose (down)-sampling as a way to adjust for imbalance and demonstrate its use for production data, where we expect an imbalance due to the controlled settings.

In the following, we first discuss our three proposed sampling strategies to select a balanced training data set. Subsequently, we present our experimental setup and methods and the production data used for our experiments. Finally, we describe our results and discuss our findings and their perspectives.

\section{Sampling approaches}
\label{sec:Approach}
The main idea of this research is to obtain a more balanced data set than the original one by sampling a more balanced training data set. We will refer to the resulting data set as the new data set. We investigate three different sampling methods, where two of them, methods (a) and (b), are based on random sampling, and the last one, method (c), is density based. 

Our random sampling methods (a) and (b) combine a unique sampling approach (i) with random sampling of the observations from the training set (ii). The idea is that approach (i) mainly samples points on the edge of the data manifold (typically low-density areas) whereas approach (ii) mainly samples points in high-density areas of the manifold. Thus combining (i) and (ii), such that the new data set consists of a 50/50 combination of samples from (i) and (ii), the new data set has an almost equal amount of data from high- and low-density areas and is thus more balanced than the original. 

\begin{figure}[h]
    \centering
    \begin{subfigure}[b]{0.32\textwidth}
        \includegraphics[width=\textwidth]{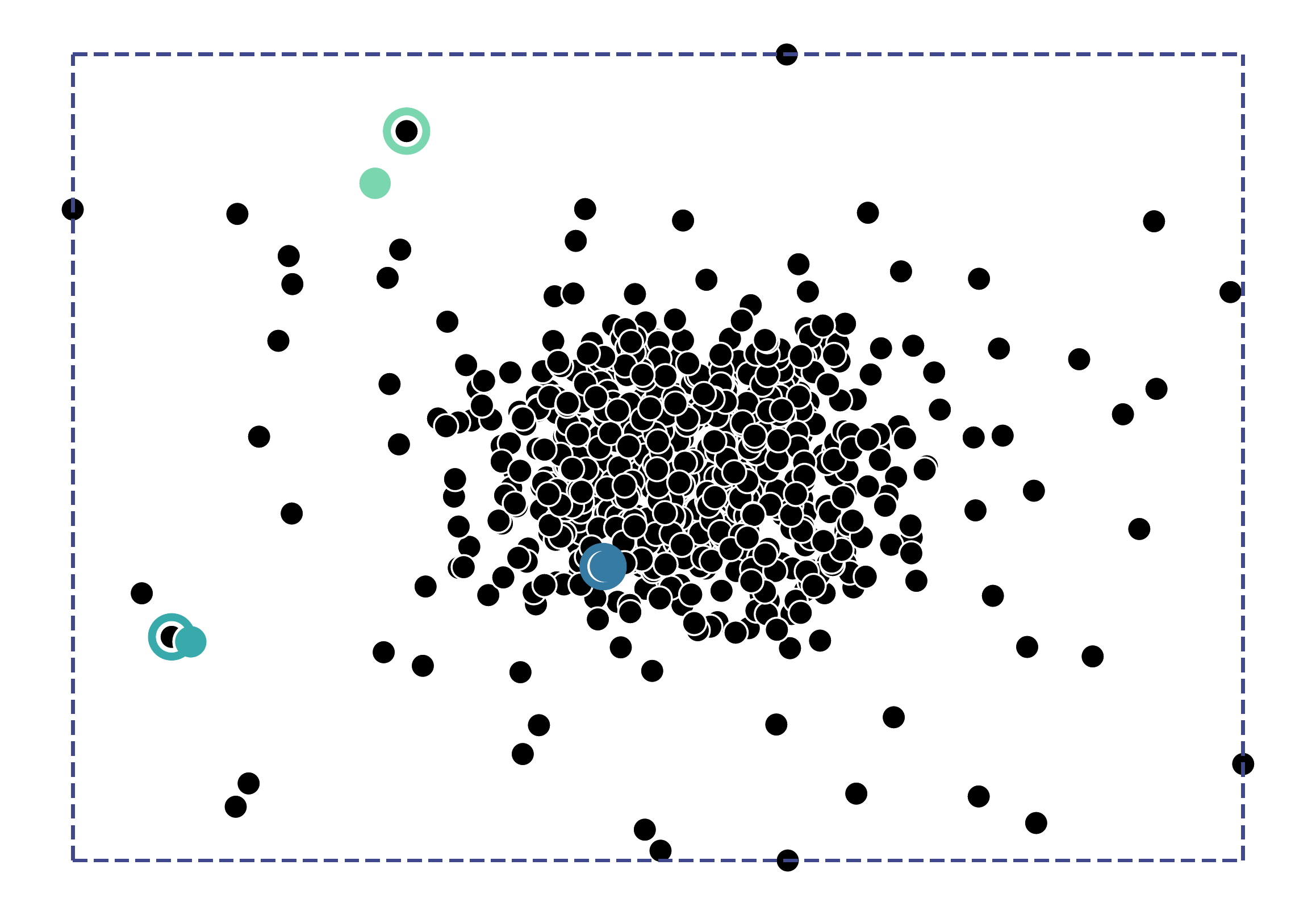}
        \caption{Random sampling a-i  ($1point$).}
        \label{fig:sampling_random_1point}
    \end{subfigure}
    ~ 
    \begin{subfigure}[b]{0.32\textwidth}
        \includegraphics[width=\textwidth]{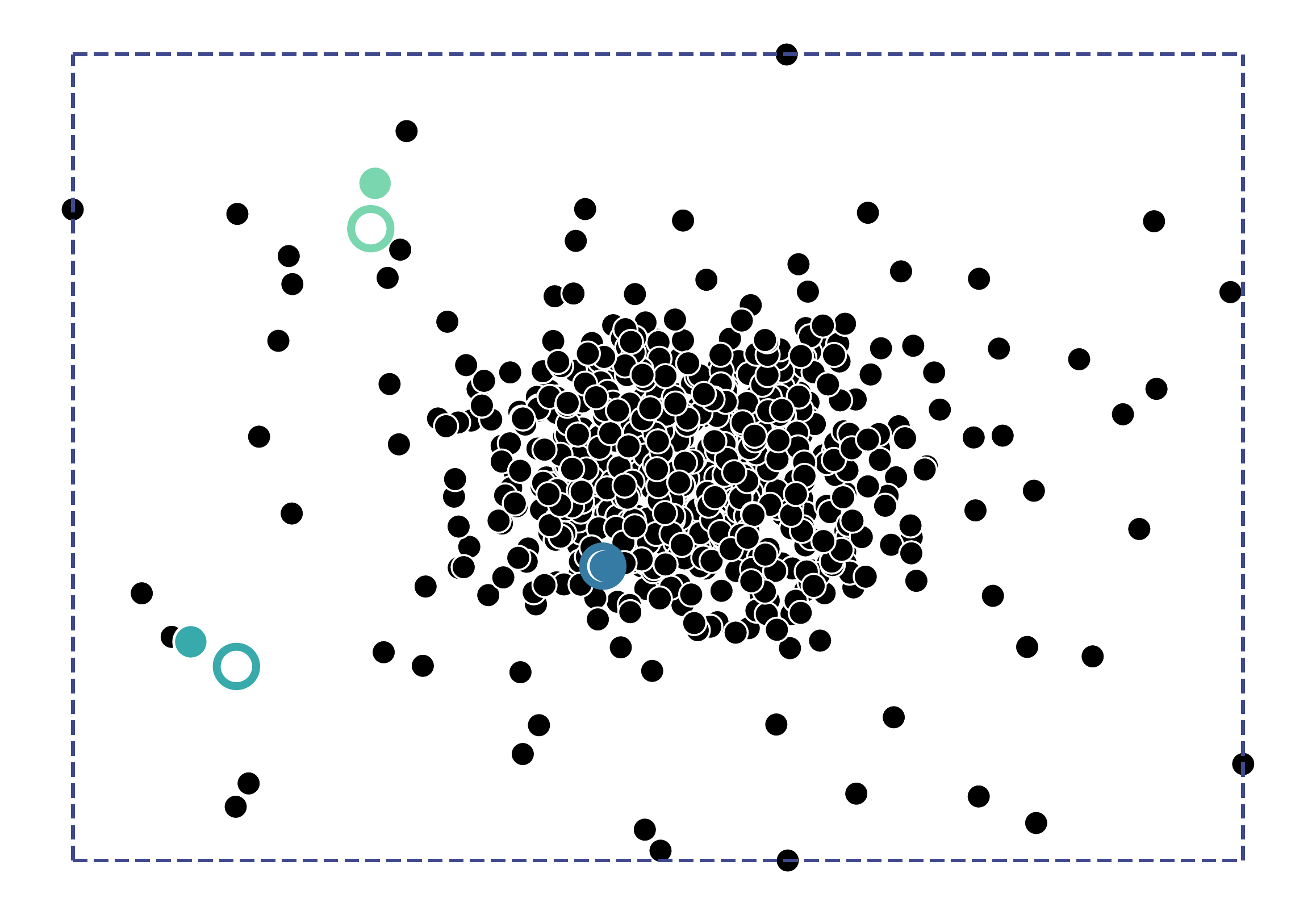}
        \caption{Random sampling b-i ($mean$).}
        \label{fig:sampling_random_mean}
    \end{subfigure}
    ~
    \begin{subfigure}[b]{0.32\textwidth}
        \includegraphics[width=\textwidth]{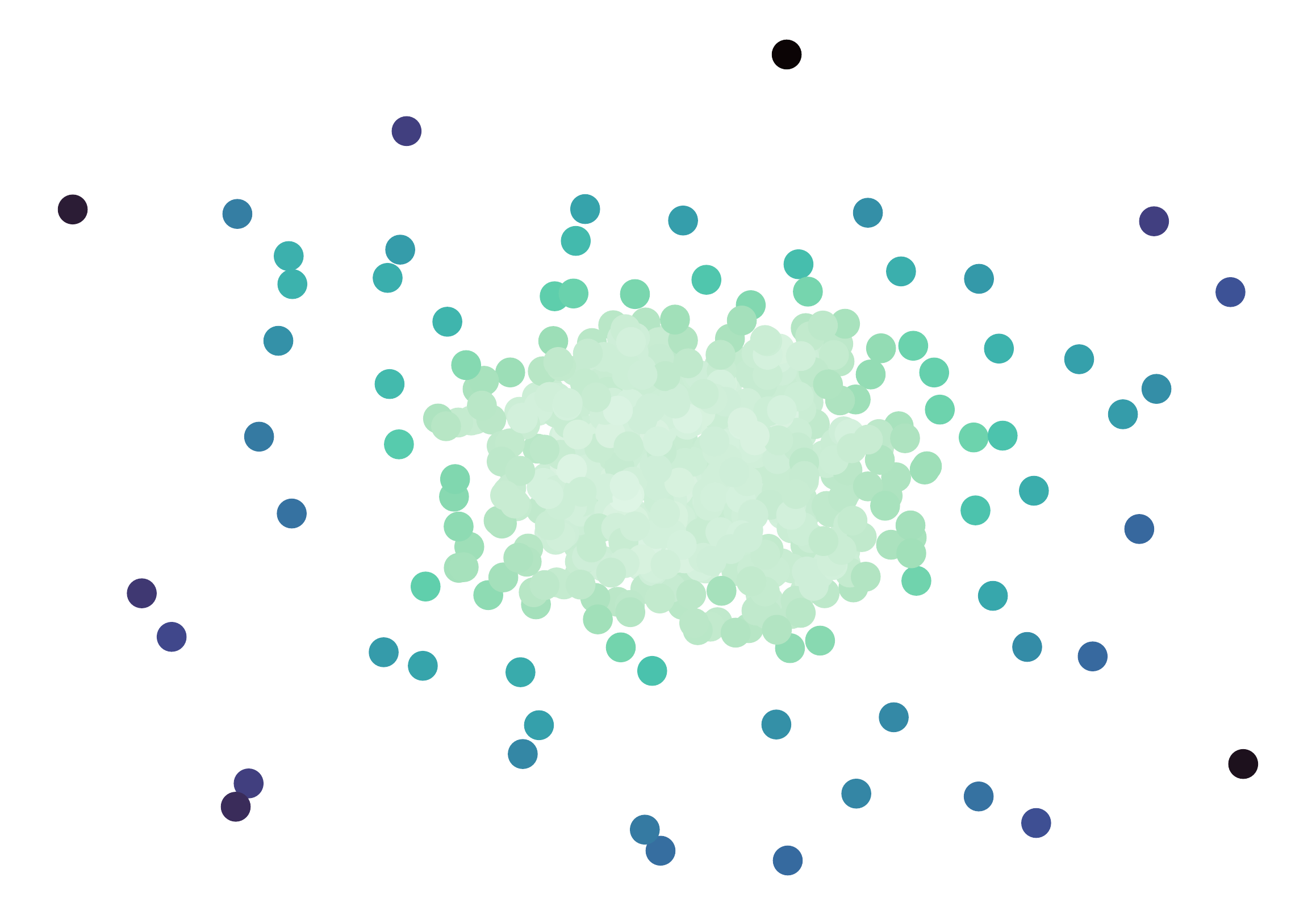}
      \caption{Density based sampling.}\label{fig:sampling_density}
    \end{subfigure}
    \caption{Illustration of the sampling methods. (a) and (b) illustrate approach (i) of random sampling. The coloured filled circles are the sampled points $z$, while the coloured rings are the (a) nearest neighbour to the sampled points or (b) the mean of the 5 nearest neighbours to the sampled points. The dotted lines represent the hyper-rectangle. (c) illustrates the density based sampling method. The colours reflect the sampling weights; the scaled mean distance to the 100 nearest neighbours. }\label{fig:sampling}
\end{figure}

Approach (i) samples points, $z$, uniformly within the hyper-rectangle spanned by the data. The sides of the hyper-rectangle are determined by the minimum and maximum value of each input variable, $x_i$, such that it has dimensions $\mathcal{Z} = [\min(x_1),\max(x_1)] \times ... \times [\min(x_{p}),\max(x_{p})]$, where $p$ is the number of variables. We then either use strategy (a), the nearest neighbour to the points $z$, denoted $1point$, or strategy (b), the mean of the 5 nearest neighbours to the points $z$, denoted $mean$, as samples in the new data set. The targets for the samples of $mean$ will be the mean of the targets for the 5 nearest neighbours. For some types of data, the mean is not necessarily meaningful, and a median approach would be a feasible alternative. Illustrations of the methods are found in Figure \ref{fig:sampling}a and  \ref{fig:sampling}b. The filled coloured circles are the sampled points $z$, while the coloured rings are the (a) nearest neighbour to the sampled points or (b) the mean of the 5 nearest neighbours to the sampled points. The dotted lines represent the hyper-rectangle, within which we sample. Since the majority of data in imbalanced data sets are concentrated on a small part of the data manifold, the nearest neighbours to most points in the hyper-rectangle will lie on the edge of the data manifold. Thus, sampling methods (a-i) and (b-i) result in a lot of samples on the edge of the data manifold.

Approach (ii) samples points randomly with equal weight from the original data set. Due to the imbalance of the data, most of the points sampled by this approach lie in high-density areas. We sample points from both (i) and (ii) corresponding to 10\% of the original (training) data. Thus, the size of the new data set for strategy (a) and (b) corresponds to 20\% of the size of the original (training) data set. 

The idea of the density based sampling method (c) is to obtain a more balanced data set by drawing a weighted random sample of the original data set with weights that reflect the inverse data density around each point. If a point is in a low-density area, the probability of drawing this point should be large, whereas if a point is in a high-density area the probability of drawing this point should be correspondingly low. We measure the data density around a point, $x$, as the mean distance to the 100 nearest neighbours of $x$. The sampling probabilities are then the mean distances scaled to sum to 1. The size of the new data set is 10\% of the original data and the sample is drawn with replacement. Figure \ref{fig:sampling_density} illustrates how the density based sampling works. The colours reflect the sampling weights and thereby the measured data density around each point. 

\section{Method and data}
We investigate the three sampling approaches by applying them on a large biopharmaceutical data set from an advanced simulation of penicillin production in a 100,000 litre penicillin fermentation system known as industrial penicillin simulation (IndPenSim) (\cite{Pen_Simulation1}, \cite{Pen_Simulation2}). The data consist of 100 batches, where the first 90 are controlled with three different production control methods, and the last 10 batches contain faults resulting in process deviations. The latter batches are often few in historical data, but also those that give insights to the dynamics of the process away from the controlled settings.

The data set contains 113,935 observations of 2,238 variables. Of these variables, 39 are process variables of which one is the penicillin concentration. The remaining 2,199 are Raman spectroscopy measurements. We disregard the Raman spectra, 5 process variables containing missing values and two with no variation and analyse the rest (31 input variables) with the goal of predicting the penicillin concentration in the tank at each observation. We hold out 20\% of the data for testing and consider the remaining  80\% for training. We compare a linear regression model trained using all of the training data to models trained using only a sample of the training data. 

\section{Results}
\label{sec:results}
\begin{figure}[h]
    \centering
    \begin{subfigure}[b]{0.49\textwidth}
        \includegraphics[width=1.05\textwidth]{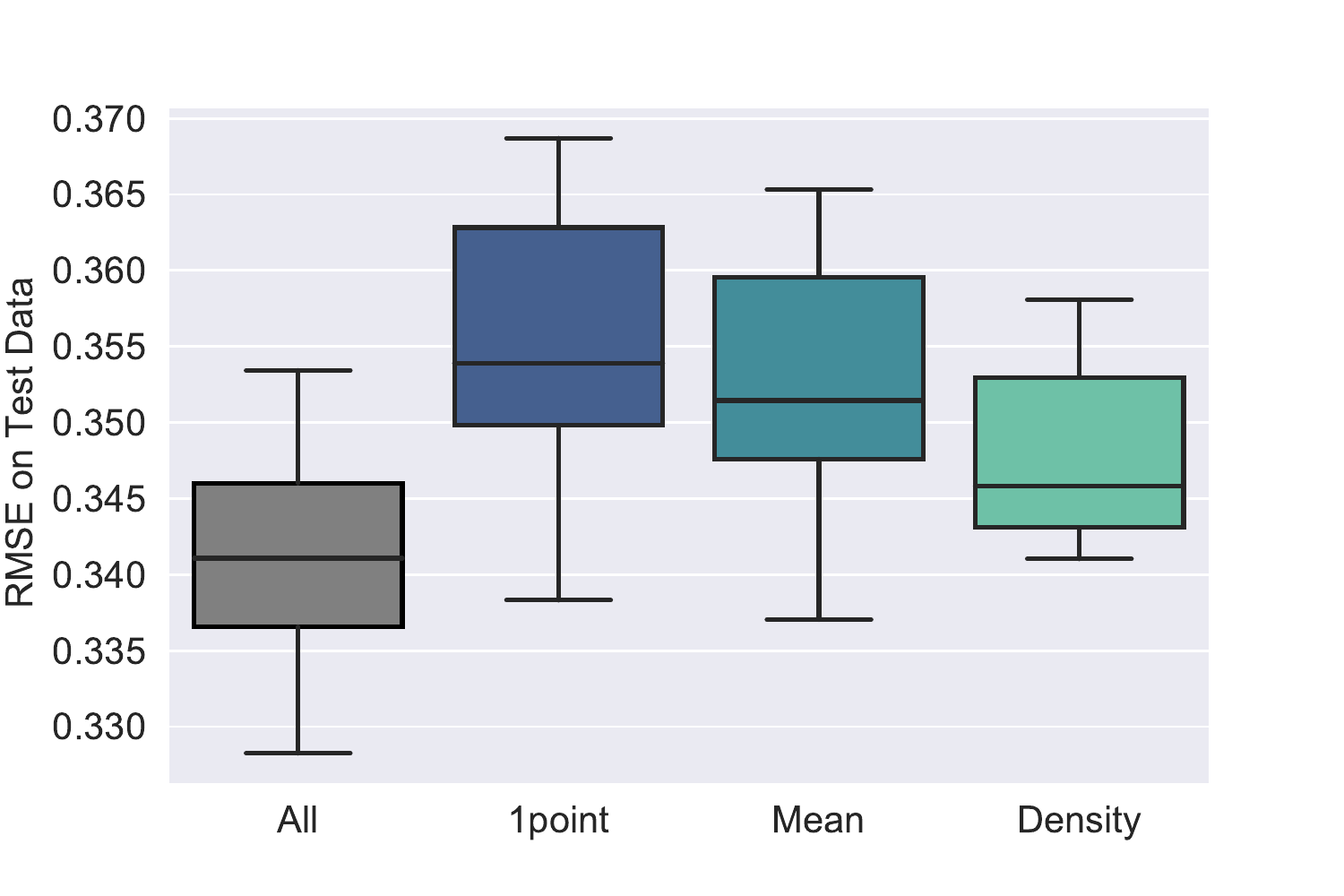}
        \caption{Full imbalanced test set.}
        \label{fig:RMSE_full}
    \end{subfigure}
    ~ 
    \begin{subfigure}[b]{0.49\textwidth}
        \includegraphics[width=1.05\textwidth]{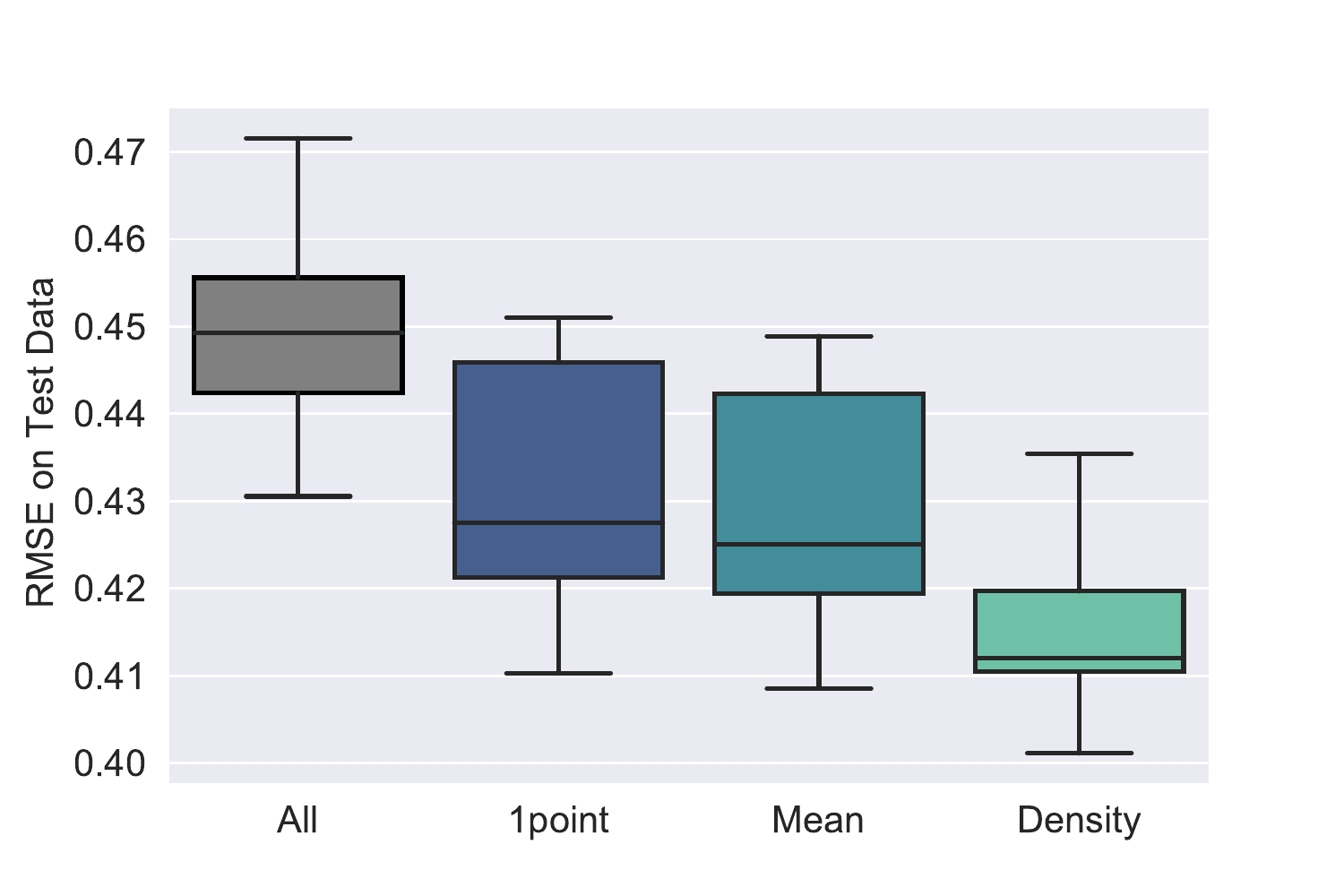}
        \caption{10\% most underrepresented observations.}
        \label{fig:RMSE_threshold}
    \end{subfigure}
    \caption{Boxplots of the RMSE on the test data after 10 iterations of fitting the linear model using either the entire training set or samples from the three sampling approaches. (a) shows the performance on the full imbalanced test set, while (b) shows the performance on the 10\% most underrepresented observations measured by the mean distance to the 100 nearest neighbours.}\label{fig:RMSE}
\end{figure}

The root mean squared errors (RMSE) of the penicillin concentrations are shown in Figure \ref{fig:RMSE}. Of the sampling approaches, the density based approach gives the lowest RMSE on average. Figure \ref{fig:RMSE}a shows the RMSE on the full test set. Since the test data is imbalanced, none of the sampling approaches improve the RMSE over using all of the training data. However, the reduced performance has a low effect size; approximately a 3\% decrease. Figure \ref{fig:RMSE}b shows the RMSE on the 10\% most underrepresented observations from the test set measured by the mean distance to the 100 nearest neighbours. Here all sampling approaches improve the RMSE over using all training data. 

\begin{figure}[h]
    \centering
    \begin{subfigure}[b]{0.49\textwidth}
        \includegraphics[width=1.1\textwidth]{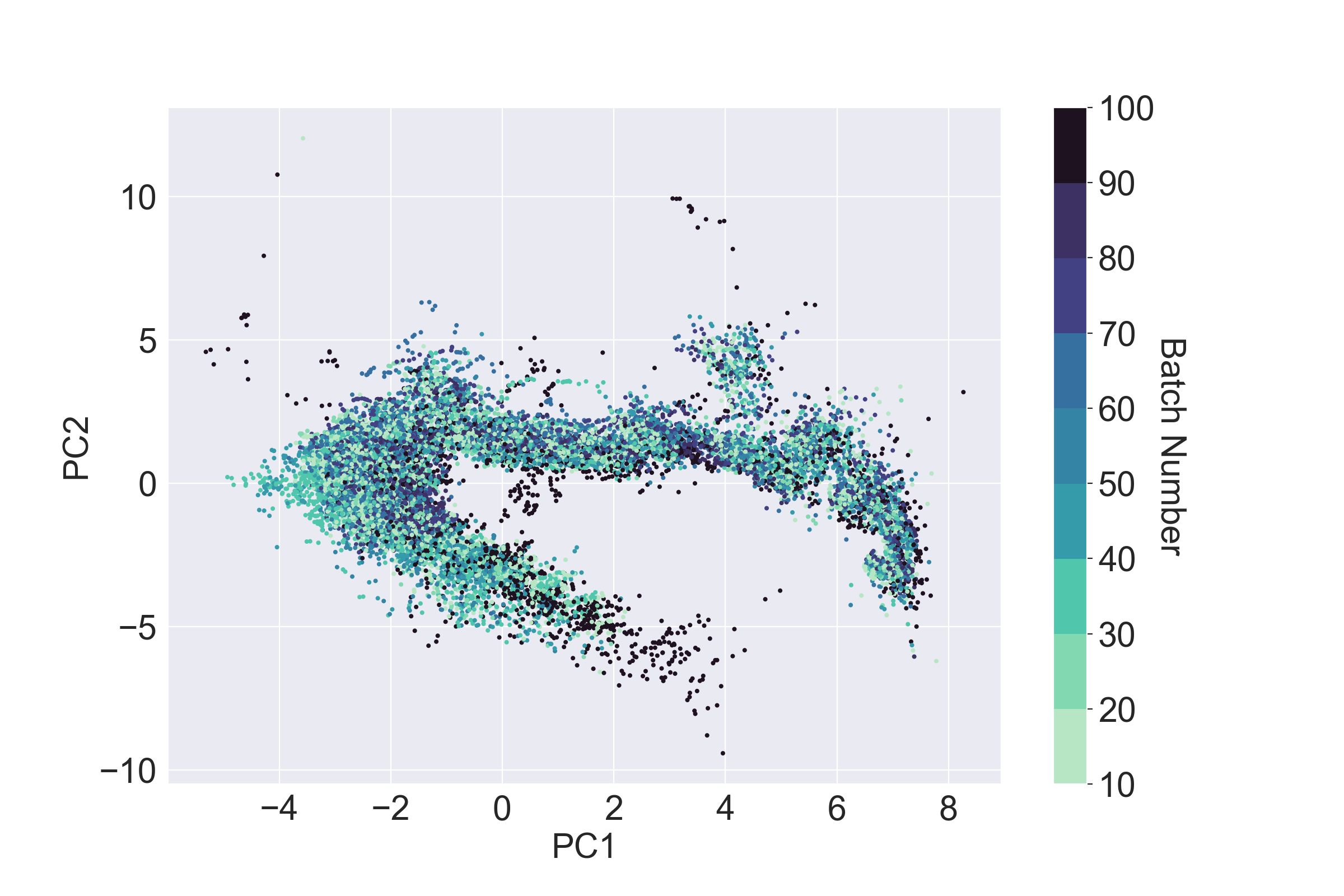}
        \caption{Batch number}
        \label{fig:PCA_batch12}
    \end{subfigure}
    \begin{subfigure}[b]{0.49\textwidth}
        \includegraphics[width=1.1\textwidth]{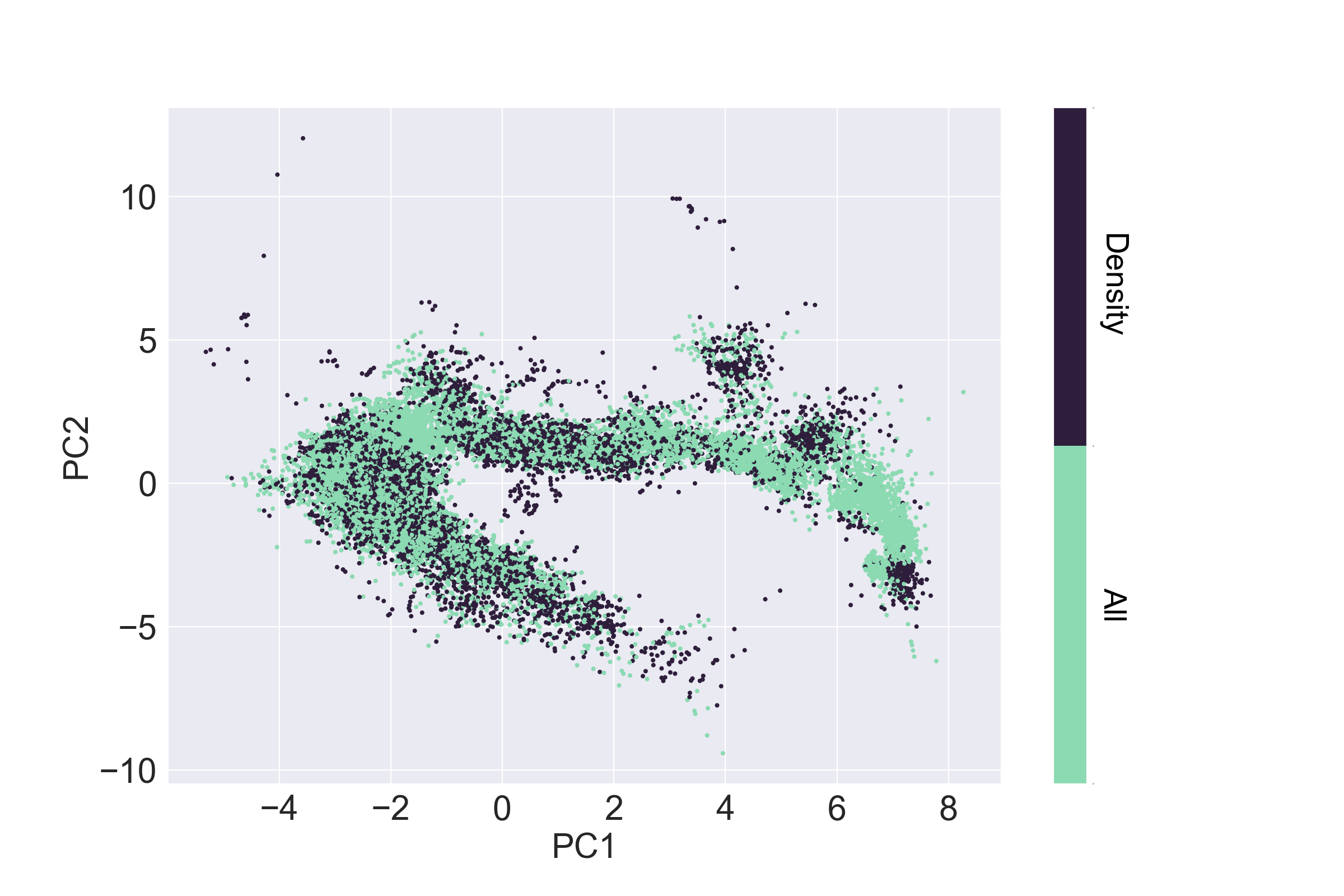}
        \caption{Residuals}
        \label{fig:PCA_res12}
    \end{subfigure}
    \caption{The test data on the first two principal components. (a) shows the data coloured according to batch number. (b) shows the data coloured according to which approach between the density sampling method and using all data gives the lowest absolute residual.}\label{fig:PCA12}
\end{figure}

Figure \ref{fig:PCA12} shows the test set observations projected onto the first two principal components, which respectively explain 29.3\% and 11.5\% of the variance. This projection illustrates how the majority of the observations lie centralised on the data manifold in high-density areas, with only few observations lying on the edges of the manifold. Figure \ref{fig:PCA12}a displays the test set observations coloured according to batch number, which shows how the majority of observations in low-density regions originate from batches with process deviations (batches 91-100). Figure \ref{fig:PCA12}b illustrates the performance difference on the test set observations projected onto the first two principal components. Black data points indicate observations where the absolute residual from the density sampling approach is smaller than the absolute residual from using all data. The sampling has improved the performance for the majority of observations on the edge of the manifold (low-density, underrepresented areas). This is particularly the case in the upper part of the figure, where residuals for observations from the lowest density regions are all improved when using the density sample to train the model over using all data.

\begin{figure}[h]
    \centering
    \begin{subfigure}[b]{0.49\textwidth}
        \includegraphics[width=1.08\textwidth]{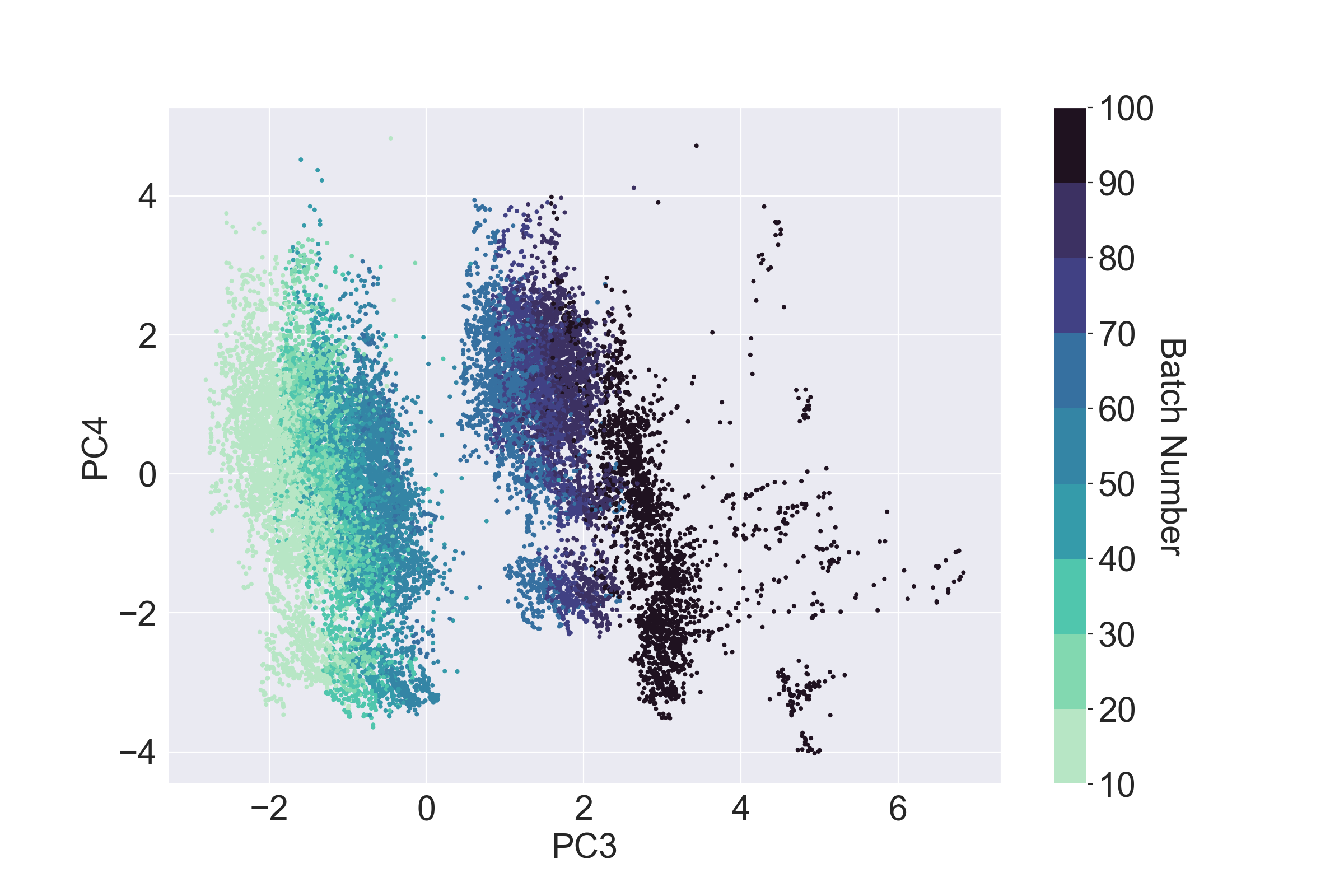}
        \caption{Batch number}
        \label{fig:PCA_batch34}
    \end{subfigure}
    \begin{subfigure}[b]{0.49\textwidth}
        \includegraphics[width=1.08\textwidth]{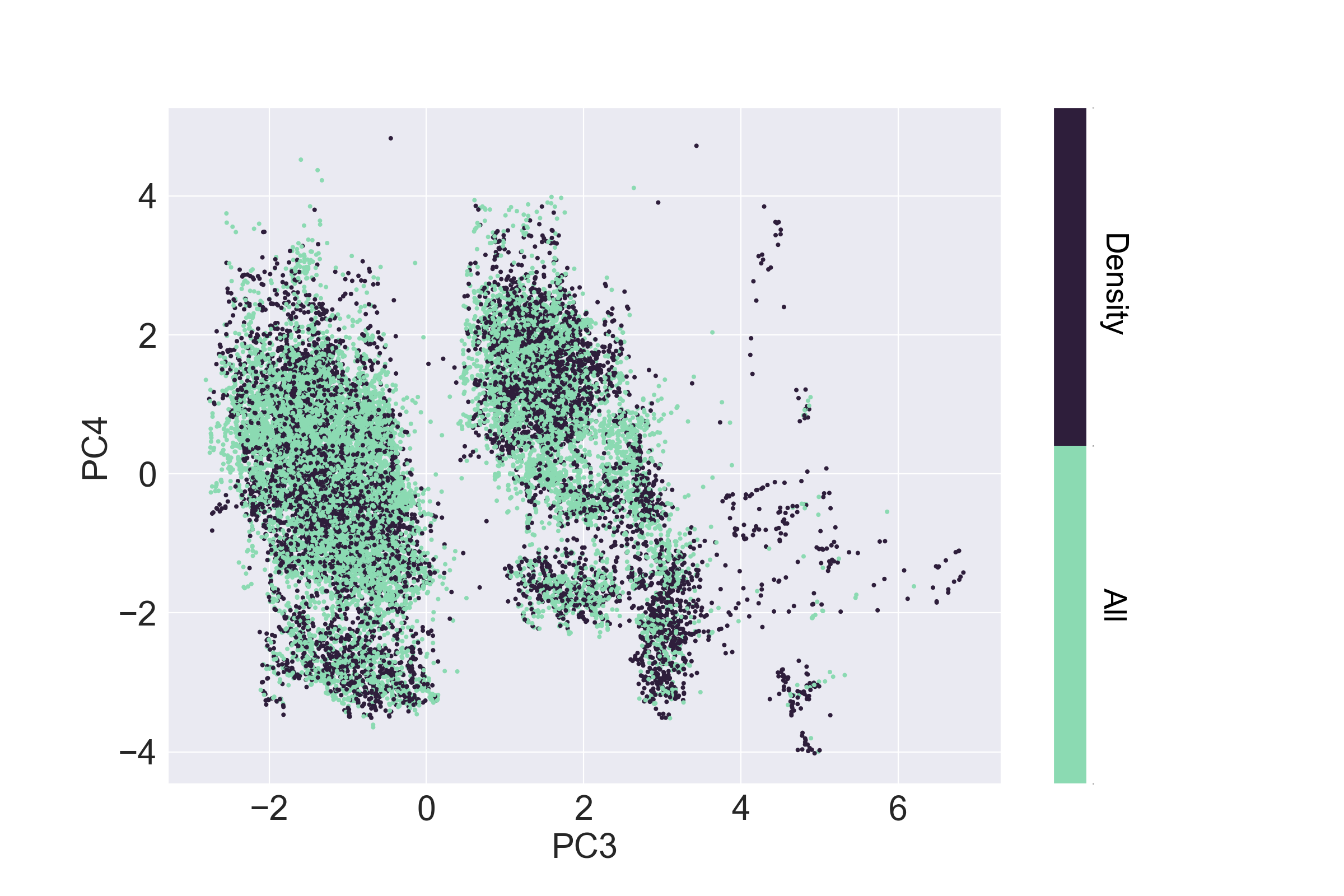}
        \caption{Residuals}
        \label{Fig:PCA_res34}
    \end{subfigure}
    \caption{The test data on principal components three and four. (a) shows the data coloured according to batch number, while (b) is coloured according to the approach with the lowest absolute residual.}\label{fig:PCA34}
\end{figure}

Figure \ref{fig:PCA34} shows similar results with the test set observations projected onto the third and fourth principal components, which explain 9.1\% and 6.9\% of the variance. Figure \ref{fig:PCA34}a shows how the third principal component captures the variation across batches, with batches 91-100 again occupying the lowest density regions. Figure  \ref{fig:PCA34}b illustrates the performance difference on the test set observations from using either all training data or the sample from the density approach. Again, the sampling has improved the performance for the underrepresented observations lying in low-density regions.

\section{Discussion}
\label{sec:discussion}
The three strategies for sampling training data to adjust for imbalance all deteriorate the overall predictive performance compared to fitting a model on all the training samples, but only with a small effect size. However, residuals for underrepresented data have improved, illustrating that sampling can drive value for underrepresented data points/areas. In this context, we would like to raise the question of how to make a \emph{balanced} and \emph{fair} evaluation, as the RMSE on imbalanced test data favours overrepresented inputs. 

While we have shown our methods apply on production data, we expect them to also apply to other types of data, where balanced representative training data could be of particular importance. This could have a potential broader societal impact on domains with historical data containing underrepresented minorities.

\bibliography{refs}

\end{document}